\documentclass{article}

\usepackage{arxiv}
\usepackage[numbers,sort&compress]{natbib}

\usepackage[utf8]{inputenc} 
\usepackage[T1]{fontenc}    
\usepackage{hyperref}       
\usepackage{url}   
\usepackage{algorithm}
\usepackage{booktabs}       
\usepackage{amsfonts}       
\usepackage{nicefrac}       
\usepackage{microtype}      
\usepackage{lipsum}
\usepackage{graphicx}
\usepackage{amsmath} 
\usepackage{amssymb} 
\usepackage{caption}
\usepackage{pgfplots}
\usepackage{tikz}
\usepackage{enumitem}
\usetikzlibrary{arrows.meta, positioning}
\usetikzlibrary{shapes.geometric, arrows.meta, positioning}

\usetikzlibrary{positioning, calc,decorations.pathreplacing}

\usepackage{pgfplotstable}
\pgfplotsset{compat=1.18}

\usepackage[noend]{algpseudocode} 

\usepackage[utf8]{inputenc}

\usepackage{algpseudocode} 

\graphicspath{ {./images/} }

\title{Isolated Bangla Handwritten Character Classification using Transfer Learning \\[0.5em]
\large Proceedings of the 2nd International Conference on Computing Advancements 2022}

\author{
\textbf{Md.\ Abdul Karim} \\
Department of Computer Science and Engineering \\
University of Asia Pacific \\
\texttt{karim.cse007@gmail.com}
\and
\textbf{S M Rafiuddin} \\
Department of Computer Science and Engineering \\
University of Asia Pacific \\
\texttt{rifat.cse@uap-bd.edu}
\and
\textbf{Md.\ Jahidul Islam Razin} \\
Department of Computer Science and Engineering \\
University of Asia Pacific \\
\texttt{razin.cse@gmail.com}
\and
\textbf{Tahira Alam} \\
Department of Computer Science and Engineering \\
University of Asia Pacific \\
\texttt{tahira.cse@example.com}
}

\begin{document}
\maketitle

\begin{abstract}
Bangla language consists of fifty distinct characters and many compound characters to be named. Several notable studies have been
performed to recognize the Bangla characters both as handwritten
and optical characters. Our approach is to use transfer learning
to classify the basic, distinct as well as compound Bangla Handwritten characters, avoiding the vanishing gradient problem. Deep
Neural Network techniques such as 3D Convolutional Neural Network (3DCNN), Residual Neural Network (ResNet), and MobileNet
have been applied to generate an end-to-end classification of all
the possible standard formations of the handwritten characters in
Bangla language. Bangla Lekha Isolated dataset is used to apply
this classification model, which has a total of 1,66,105 Bangla Character images sample data categorized in 84 distinct classes. This
classification model achieved 99.82\% accuracy on training data and
99.46\% accuracy on test data. Comparison has been made among
The various state-of-the-art benchmarks of Bangla Handwritten
Characters classification, which shows that this proposed model
got better accuracy in classifying the data.
\end{abstract}

\noindent\textbf{Keywords—} Deep learning; Convolution Neural Network; Transfer Learning; Bangla handwriting recognition; MobileNet; Residual Neural Network.

\section{Introduction}
\label{sec:introduction}

Within all the languages spoken within the Indian Subcontinent,
Bangla, also known as Bengali, is the second most common language. With 242 million people from around the world, Bangla is the
sixth most common language in the world. Bangla is the National
Language in Bangladesh \cite{simons2017ethnologue}. In these developing countries like
India and Bangladesh, people use handwritten documents. This is
the fundamental medium of communication in the case of documentation. A student using a pen to make a note in a lecture is a clear
example of this. In many educational organizations and institutions,
the use of handwritten notices and circulars in national dialects is
still preferred over printed ones. Therefore, handwritten character
recognition is of great importance for the Bangla language.

\begin{figure}[htbp]
    \centering
    \includegraphics[width=0.5\textwidth]{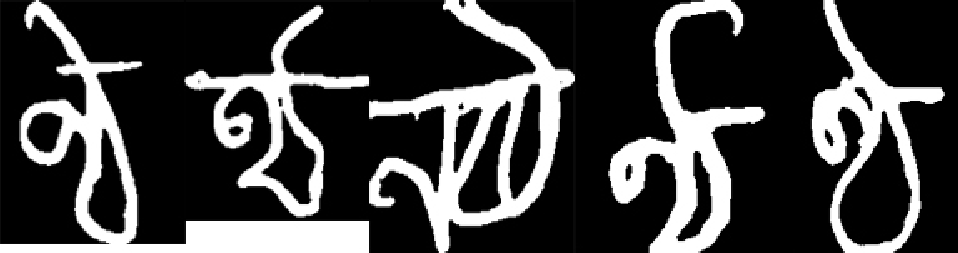} 
    \caption{Depending on the writers the same letter has a different form.}
    \label{fig:bangla-variation}
\end{figure}

Automatic identification of handwriting has practical usage in
libraries, postal services, banks, etc. A huge number of handwritten
documents are processed by such organizations daily. Only half
of these documents can be recognized \cite{pal2012handwriting}. An automated handwriting recognizer can solve this problem and also save a lot of
time. The development of deep learning technology has made it
possible for several complicated large-scale identification tasks,
particularly at ImageNet’s Large Visual Recognition Challenge \cite{russakovsky2015imagenet}.
Deep learning architects including AlexNet \cite{krizhevsky2012imagenet}, VGGNet \cite{simonyan2014very} and
ResNet \cite{he2016deep} have already demonstrated their ability to provide highrecognition efficiency on large-scale image classification methods.
The AlexNet is the pioneer in restoring the large-scale classification
of neural network machinery by transformation of the initial CNN
\cite{lecun1998gradient}. VGGNet then reinforced the concept of CNN by introducing
simpler filters with a longer architecture and the winner of ILSVRC
2014. By introducing CNN, VGGNet strengthened the notion of
simpler, lengthier-architecture filters. Following that trend, ResNet
used an excellent architecture to construct the network for the 2016
ImageNet competition using a modular design. Although ResNet
uses a much deeper architecture than VGGNet-16 and VGGNet-19,
with a smaller number of parameters, the architecture converges
quicker. While deep learning dominates in various research areas
including its strong discrimination capabilities. MobileNet \cite{su2018mobilenet,li2018mobilenetssd}
is a new CNN model that provides satisfactory classification accuracy while it reduces the number of parameters in comparison to
the conventional convolution (Conv) layer in CNN models.

Bangla Handwritten Character Recognition is considered a difficult task. In this article, to recognize complex Bangla Handwritten
Characters, three major issues are considered. Those are-
Recognition of convoluted edges.• Differentiate by repeating the same pattern in different characters.• Different patterns that are written with the same character.
Characters or digits handwritten in Bangla seldom have defined
measurements. It is normal for the two characters to consider different geometric structures. Many of the characters have rounded
corners, e.g. the handwritten structures of characters "SHa" and
"za" are rather conflicted geometrical structures. Second, several
different letters in alphabets in Bangla have the same pattern repeated. These structures of characters make the task of classification
more difficult. Lastly, writing style varies from person to person and the character’s geometrical structure fluctuates from sizes and
angles, for example in Fig.~\ref{fig:bangla-variation} depending on the writers the same
letter has a different form.

In this paper, we propose a method to
resolve the above challenges in three modified models of 3DCNN,
ResNet50, and MobileNet to solve HCR for Bangla characters. The
contributions of this paper can be described in two folds:
1) We propose the modified architectures of 3DCNN, ResNet50,
and MobileNet, capable of end-to-end learning and modernized
classification on relatively large data systems.
2) To fix the issue of Bangla HCR, which can be used as the basis
for comparison in the future, we provide a quantitative overview
of the results of several state-of-the-art deep learning models.

The remainder of the document is structured accordingly as follows:  

\begin{itemize}
    \item In Section~\ref{sec:related}, Some related research is presented to gain insight into the research we have proposed. 
    \item The proposed method is explained in Section~\ref{sec:method}. 
    \item Implementation Tools is explained in Section~\ref{sec:tools}. 
    \item In Section~\ref{sec:results}, Our research findings and contrast with other approaches have also been shown. 
    \item In Section~\ref{sec:comparison}, we compare our model with other research papers. 
    \item Finally, in Section~\ref{sec:conclusion}, We are giving our concluding and discussing our hopes for the future.
\end{itemize}

\section{Related Works}
\label{sec:related}

T.K. Bhowmik, U. Bhattacharya, and Swapan K. Parui proposed a
model MLP Classifier Based on Stroke Features \cite{bhowmik2004recognition}. For this, they are
using their own created Bangla basic characters dataset where there
is no isolated dataset. This dataset has 25,000 collected samples of
image data from different people of West Bengal in India. From
this, they used 17,500 data for training and 3,000 and 4,500 data for
testing purposes. They tested this dataset using different numbers
of hidden nodes and they got maximum accuracy for 110 hidden
nodes. Their training accuracy is 89.43\% and their testing accuracy
is 84.33\%.

U. Bhattacharya, M. Shridhar, and S.K. Parui proposed a model
using Multi-layer perceptrons and Back-propagation \cite{bhattacharya2006recognition}. For this,
they used a dataset that has 20,187 Bangla handwriting characters
and they classified these data into 50 different classes. From this,
they used 10,000 data for training purposes and the rest of the
10,187 data for testing purposes. After training and testing, their
maximum training accuracy is 94.65\% and the testing accuracy is
92.14\%.

Parui et al. \cite{parui2008online} proposed a model using a dataset which has
24,500 data sample written by 70 people. This dataset has 50 different classes. From this, they took 14,000 data for training purposes,
7,000 and 3,500 data for testing purposes. Using the Hidden Markov
model, after training and testing they got 87.7\% accuracy on testing
data.

Bhowmik et al. \cite{bhowmik2009svm} proposed a model which is Support Vector
Machine \cite{suykens1999ls} based hierarchical classification schemes to detect
Bangla handwriting. They used a dataset that has 45 different classes
and the whole dataset sample size is 27,000. They divided this
dataset into three different sets, which are the training set, testing
set, and validation set. Using this dataset their proposed model got
an accuracy of 89.22\%.

Using Multi-layer Perceptron \cite{pal1992mlp} and Support Vector Machine
\cite{suykens1999ls} Mohiuddin, Sk and Bhattacharya proposed a model \cite{mohiuddin2011unconstrained}. They
used two different datasets. One dataset has 14,073 sample data
corresponding to 50 cities in Bangla, which is written by 163 writers
and other datasets have 27,798 data corresponding to 110 cities in
Bangla, which is written by 298 writers. The training purpose takes
7,757 data from 50 cities dataset which are written by 87 writers and
take 11,000 sample data from the 110 city datasets. For the testing
purpose, they took 6,516 sample data which is written by 76 writers
from the 50 cities datasets and takes 16,798 sample data which is
written by 161 writers from the 110 city datasets. This proposed
method was implemented on this dataset and they got an accuracy
of 88.79\% for the 50 cities dataset and 87.20\% accuracy for the 110
cities dataset.

Alif, Mujadded Al Rabbani, and Ahmed proposed a modified
version of the ResNet-18 model for recognization of Bangla isolated
handwriting \cite{alif2017isolated}. For this research work, they used two different
large datasets which are the BanglaLekha-Isolated dataset \cite{biswas2017banglalekha} and
CMA-TERdb dataset \cite{das2014benchmark}. CMA-TERdb dataset has 231 classes of
image and the BanglaLekha-Isolated dataset has 84 classes. These
classes included vowel, consonant, and conjunct character images.
From this, they only took 132,884 data for training sets and 33,221
data for testing sets. After training and testing, they got 95.10\%
accuracy for testing.

\section{Methodology}
\label{sec:method}

We will explain our system for detecting and classifying isolated
Bangla handwriting from the ‘BanglaLekha-Isolated’ image dataset
in this segment. Our approach model is split down into two phases.

\begin{itemize}
    \item \textbf{Input Image Processing}

    \item \textbf{Training and Testing the Model} 

\end{itemize}

\subsection{Input Processing}
\label{sec:input}

Before using this dataset (Section~\ref{sec:dataset}) for training and testing purposes,
we need to pre-process the dataset. For convenience, this dataset
is divided into four-phase which are Vowel, Consonant, Numeric,
and Compound characters. After that, we resize this dataset image
in different sizes and different dimensions and append them in
an array using OpenCV \cite{bradski2008learning}. After that we convert this dataset as
NumPy array and using NumPy, we label this dataset according to
their class names using NumPy \cite{vanderwalt2011numpy}. Then we split them into data
and labels. This label means target class name convert vector to
binary class matrices using Keras np-utiles \cite{ketkar2017keras} and split this data
into training and testing set using Scikit-learn \cite{pedregosa2011scikit}.

\begin{figure}[htbp]
    \centering
    \includegraphics[width=0.75\textwidth]{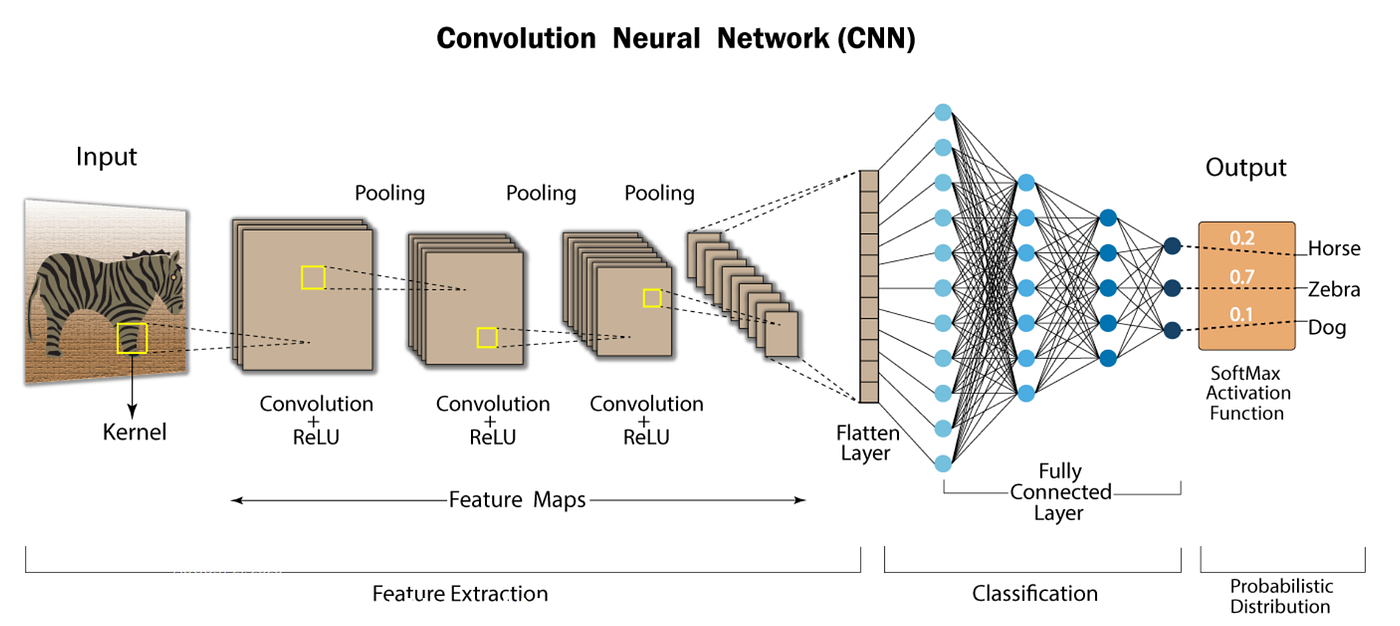}
    \caption{Convolution Neural Network Architecture.}
    \label{fig:cnn-arch}
\end{figure}

\subsection{Training and Testing the Model}
\label{sec:training}

For training and testing on the above-mentioned dataset, we evaluate
this by three different models: 3D Convolution Neural Network
(3DCNN) in Section~\ref{sec:3dcnn}, ResNet50 in Section~\ref{sec:resnet50}, and
MobileNet in Section~\ref{sec:mobilenet}.

\subsubsection{3D Convolution Neural Network (3DCNN)}
\label{sec:3dcnn}

We applied 3D-CNN for training our interface. It is one of the Convolution
Neural Network (CNN) based methods of deep learning to boost
our model’s accuracy. The architecture of CNN’s layers is distinct
from other neural networks. CNNs have various layers, such as
the input layer, convolution layer, hidden layer, pooling layers, and
output layer. Fig.~\ref{fig:cnn-arch} is showing the CNN Architecture. The primary feature of CNN is the convolution layer. This sheet will be
used to remove points, corners, and colors. It also assists with the
recognition of various forms, digits, and specific sections. For better
results, we used the 3D convolution layer in our proposed model.
On each one of them, we used a 3D convolution layer with a filter
size of 32 and a 3$\times$3 pooling layer. We have used several deep learning methods to enhance the model’s accuracy. We used Rectified
Linear Unit (ReLU) activation layer \cite{li2017convergence} to reinforce the positive
results and analyze the bad findings downwards.

Activation feature Softmax \cite{jang2016categorical} is used to generate just two
probabilistic values 0 or 1 in the classification model. Max-pooling
\cite{wang2019dominant} layer is used to reduce the number of parameters resulting
from image matrices and to only take the maximum parameter from
matrices. For the regularization of our platform, we used dropout. Finally, we used Adam optimizer \cite{tato2018adam} in training data to change
the network widths.

\begin{figure}[htbp]
    \centering
    \includegraphics[width=0.75\textwidth]{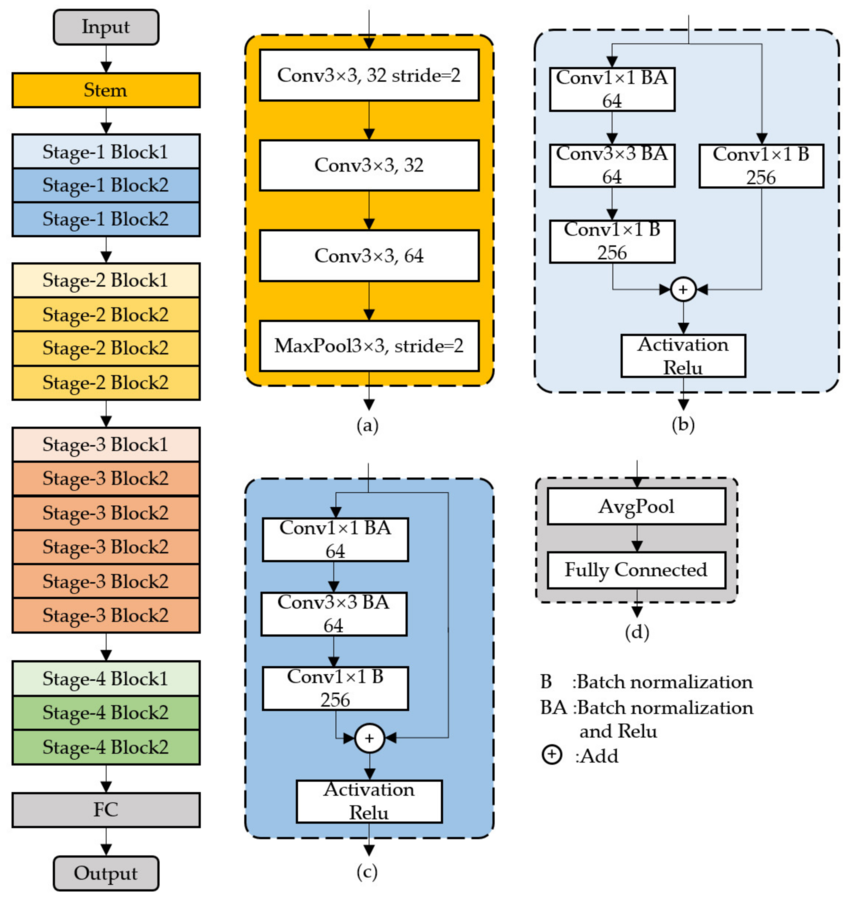}
    \caption{ResNet50 Architecture \cite{ji2019optimized}.}
    \label{fig:resnet-arch}
\end{figure}

\subsubsection{ResNet50}
\label{sec:resnet50}

Residual Networks (ResNet) is a popular neural network used by multiple Computer Vision activities as the
backbone. ResNet is just another deep network leader. The state-of-the-art Deep Neural Network had been moving deeper and deeper
before ResNet. Nevertheless, due to the infamous vanishing gradient problem, deep networks are difficult to learn, because the
gradient is back-propagated to earlier layers, repetitive multiplication will render the gradient exponentially low. By adding a skip
link to match the data from the previous layer into the next layer
without any data change, ResNet can provide up to 152 layers of a
very deep network.

We used ResNet50 for handwriting identification, which is also
used for image processing, face recognition, and even entity identification. Many approaches have been developed based on ResNet50
or ResNet101 for their simplicity and practicality. ResNet was used
in several fields including tracking, segmentation, and recognition.
Alpha Zero also uses ResNet and ResNet runs perfectly.

In our proposed model, the first seven models make trainable
false and the other model makes trainable true. We used 64 filter
size and 3$\times$3 pooling layer and also used 3$\times$3 kernel size. A kernel is
a function that helps to categorize non-linear problems. We used
the ReLU activation function and dropout 0.5 which is helpful to
regularize the data and finally, we used the Softmax activation
function to decide the acceptance threshold value. We resized the
image into 160$\times$160 pixels and train and test with a batch size of 32
with Adam optimizer. Fig.~\ref{fig:resnet-arch} shows the ResNet50 architecture.

\subsubsection{MobileNet}
\label{sec:mobilenet}

\begin{figure}[htbp]
    \centering
    \includegraphics[width=0.75\textwidth]{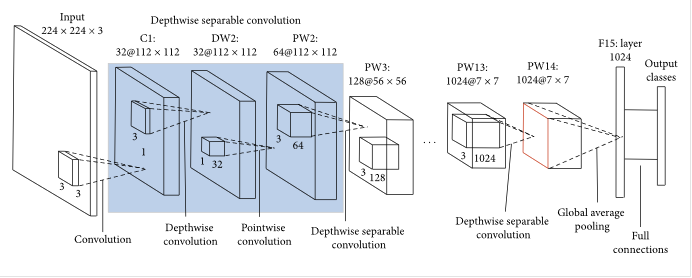}
    \caption{MobileNet Architecture \cite{su2018mobilenet}.}
    \label{fig:mobilenet-arch}
\end{figure}

MobileNet \cite{su2018mobilenet,li2018mobilenetssd} is a new CNN model aimed
at providing respectable classification accuracy with a reduced
amount of parameters compared to the conventional convolution
(Conv) layer in CNN models. That consists of a convolution depth
layer (DW Conv) and a convolution level layer. A DW Conv layer
has a kernel for each input function map channel that consists of
a kernel for each frame. Two-dimensional convolutions are performed channel-wise independently. In comparison, the PW Conv
layer is a special case of a general Conv layer and has a kernel
size of 1/1/a component N/M while a general Conv layer increasing have kernels of a more common size of K/K/N/M. There are
some general Conv layers and often DSC layers that can shape the
MobileNet models. Fig.~\ref{fig:mobilenet-arch} shows the MobileNet architecture.

Our Proposed model first imports a hand-held system interface
and discards the last 1000 layers of neurons. We added dense layers
to allow the model to learn more complex functions and differentiate
for better performance. Finally, we used the Softmax activation
function. After that, the first twenty layers of the model are made
non-trainable and the rest of them are made trainable with batch
size 32. Using Adam optimizer we trained and tested the datasets.
We trained this model with an image size between 50$\times$50 pixels
to 160$\times$160 pixels. We got a better result when the image size is
approximately 100$\times$100 pixels.

\section{Implementation Tools}
\label{sec:tools}

We used Jupyter Notebook \cite{perkel2018jupyter}, an open-source software program.
It helped to build an ecosystem to conduct our experiments and
also to generate and exchange all sorts of documentation such as
code, language, calculations, and visualizations. The CNN model is
designed using TFLearn (deep learning library), a Python package
constructed on top of TensorFlow \cite{abadi2016tensorflow}. We also used the Keras
\cite{gulli2017deep} library, which is written using Python, and TensorFlow works
on it.

\section{Result and Analysis}
\label{sec:results}

This segment addresses the interpretation of the findings coupled
with the consequence of the whole program. We broke the result
into three categories. The categories are collections of data in 
Section~\ref{sec:dataset}, experiment results analysis in 
Section~\ref{sec:analysis}, and comparison in 
Section~\ref{sec:comparison}.

\subsection{Experimental Dataset}
\label{sec:dataset}

\begin{figure}[htbp]
    \centering
    \includegraphics[width=0.75\textwidth]{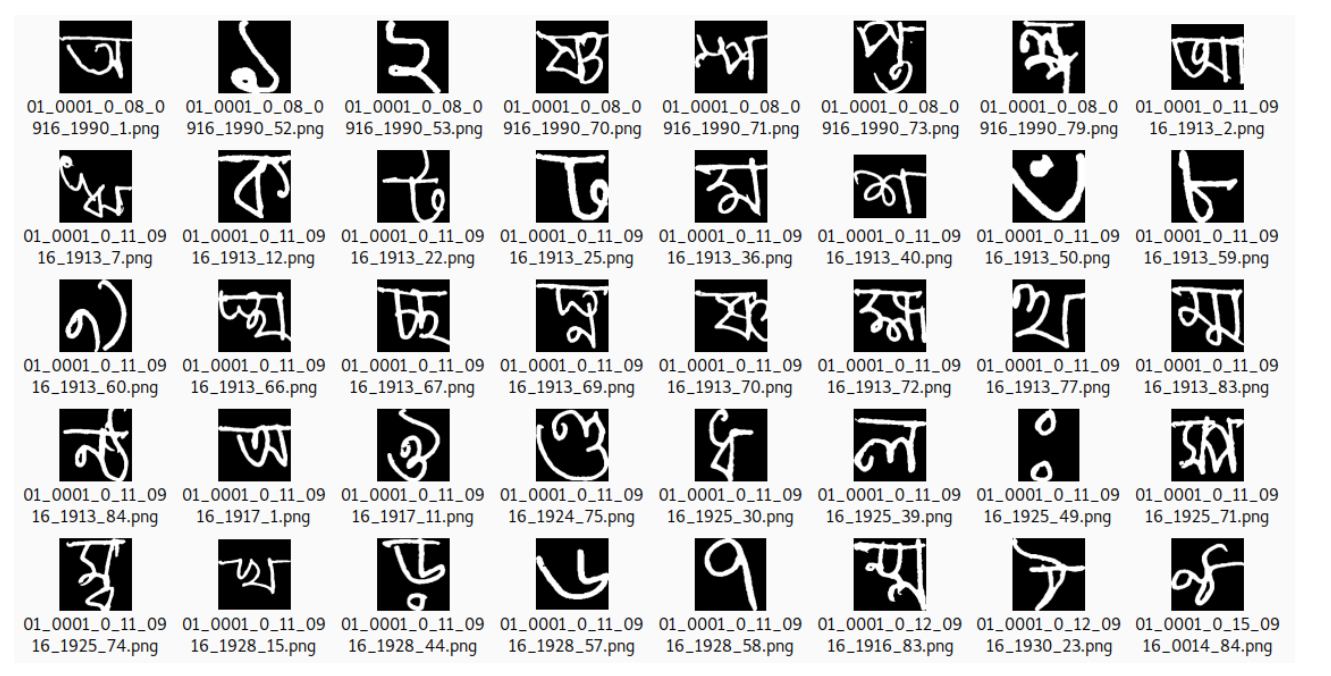}
    \caption{Sample images from the BanglaLekha-Isolated dataset.}
    \label{fig:dataset-sample}
\end{figure}

To train our model, we choose the dataset name ``BanglaLekha-Isolated'' from Menedely which is provided by Mohammed, Nabeel
\cite{biswas2017banglalekha}. This dataset has 84 different classes where 50 classes are Bangla basic characters, 10 classes are Bangla numeric and 24 classes are
Bangla compound characters. Each class has an average of 2000
sample image data and different 84 classes have a total of 166,105
images. Fig.~\ref{fig:dataset-sample} shows the dataset sample. From this dataset, we
take 80\% data for the training purpose and the rest 20\% data for the
testing purpose.

\subsubsection{Experiment Results Analysis}
\label{sec:analysis}

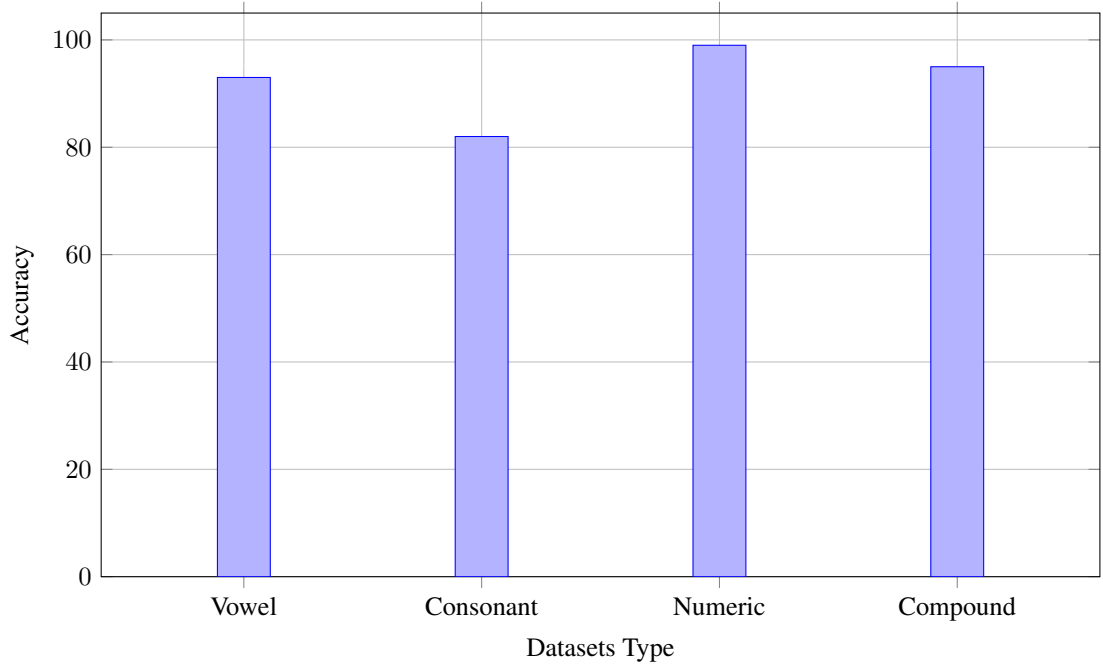
\begin{figure}[htbp]
\centering
\begin{tikzpicture}
\begin{axis}[
    ybar,
    bar width=20pt,
    width=0.9\textwidth,
    height=0.55\textwidth,
    ymin=0, ymax=105,
    ylabel={Accuracy},
    xlabel={Datasets Type},
    symbolic x coords={Vowel, Consonant, Numeric, Compound},
    xtick=data,
    nodes near coords,
    nodes near coords align={vertical},
    every node near coord/.append style={font=\bfseries\footnotesize, color=white},
    yticklabel style={/pgf/number format/fixed, /pgf/number format/precision=0},
    enlarge x limits=0.2,
    grid=major
]
\addplot coordinates {(Vowel,93) (Consonant,82) (Numeric,99) (Compound,95)};
\end{axis}
\end{tikzpicture}
\caption{3DCNN Testing accuracy graph according to datasets type.}
\label{fig:3dcnn-accuracy}
\end{figure}

To recognize handwriting on the above mentioned dataset, we evaluated
this by three different models: 3D Convolution Neural
Network (3DCNN) in Section~\ref{sec:res-3dcnn}, ResNet50 in Section~\ref{sec:res-resnet50}, 
and MobileNet in Section~\ref{sec:res-mobilenet}.

\paragraph{3D Convolution Neural Network (3DCNN).}
\label{sec:res-3dcnn}

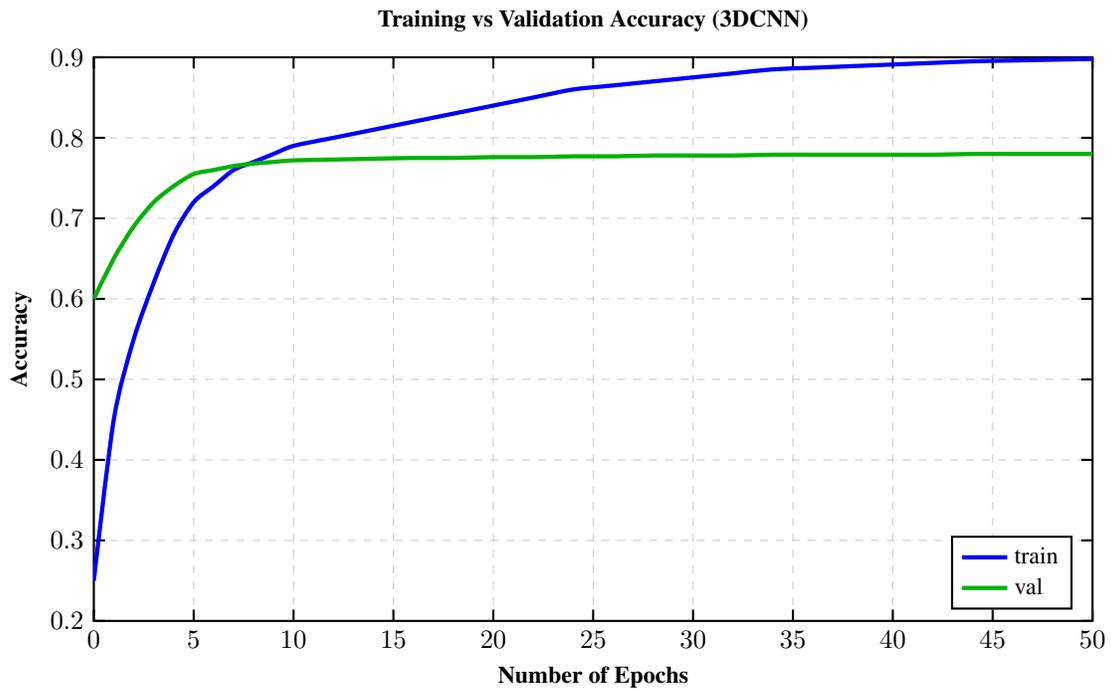
\begin{figure}[htbp]
\centering
\begin{tikzpicture}
\begin{axis}[
    width=0.9\textwidth,
    height=0.55\textwidth,
    xmin=0, xmax=50,
    ymin=0.2, ymax=0.9,
    xlabel={Number of Epochs},
    ylabel={Accuracy},
    title={Training vs Validation Accuracy (3DCNN)},
    grid=major,
    grid style={dashed,gray!40},
    tick style={black, thick},
    label style={font=\bfseries\small},
    title style={font=\bfseries\small},
    legend style={draw=black, fill=white, font=\small, at={(0.98,0.02)}, anchor=south east},
    legend cell align={left},
    thick
]

\addplot+[smooth, ultra thick, color=blue, mark=none]
coordinates {
(0,0.25) (1,0.45) (2,0.55) (3,0.62) (4,0.68) (5,0.72) (6,0.74) (7,0.76)
(8,0.77) (9,0.78) (10,0.79) (12,0.80) (14,0.81) (16,0.82) (18,0.83)
(20,0.84) (22,0.85) (24,0.86) (26,0.865) (28,0.87) (30,0.875) 
(32,0.88) (34,0.885) (36,0.887) (38,0.889) (40,0.891) (42,0.893)
(44,0.895) (46,0.896) (48,0.897) (50,0.898)
};
\addlegendentry{train}

\addplot+[smooth, ultra thick, color=green!70!black, mark=none]
coordinates {
(0,0.60) (1,0.65) (2,0.69) (3,0.72) (4,0.74) (5,0.755) (6,0.76) (7,0.765)
(8,0.768) (9,0.77) (10,0.772) (12,0.773) (14,0.774) (16,0.775)
(18,0.775) (20,0.776) (22,0.776) (24,0.777) (26,0.777) (28,0.778)
(30,0.778) (32,0.778) (34,0.779) (36,0.779) (38,0.779) (40,0.779)
(42,0.779) (44,0.780) (46,0.780) (48,0.780) (50,0.780)
};
\addlegendentry{val}

\end{axis}
\end{tikzpicture}
\caption{3DCNN Training and Testing accuracy after fifty iterations.}
\label{fig:3dcnn-train-test}
\end{figure}

\begin{figure}[htbp]
\centering
\begin{tikzpicture}
\begin{axis}[
    width=0.9\textwidth,
    height=0.55\textwidth,
    xmin=0, xmax=50,
    ymin=0.2, ymax=3.0,
    xlabel={num of Epochs},
    ylabel={loss},
    title={train\_loss vs val\_loss},
    grid=both,
    grid style={dashed,gray!45},
    tick style={black, thick},
    label style={font=\bfseries\small},
    title style={font=\bfseries\small},
    legend style={draw=black, fill=white, font=\small, at={(0.82,0.85)}, anchor=north west},
    legend cell align={left},
    thick
]

\addplot+[smooth, ultra thick, color=blue, mark=none]
coordinates {
(0,2.80) (1,1.80) (2,1.30) (3,1.10) (4,1.00) (5,0.90)
(6,0.85) (8,0.78) (10,0.70) (12,0.65) (14,0.60)
(16,0.56) (18,0.52) (20,0.50) (22,0.48) (24,0.46)
(26,0.44) (28,0.42) (30,0.40) (32,0.39) (34,0.38)
(36,0.37) (38,0.36) (40,0.35) (42,0.34) (44,0.33)
(46,0.32) (48,0.31) (50,0.35)
};
\addlegendentry{train}

\addplot+[smooth, ultra thick, color=green!70!black, mark=none]
coordinates {
(0,1.60) (1,1.20) (2,1.05) (3,0.95) (4,0.90) (5,0.88)
(6,0.87) (8,0.90) (10,0.92) (12,0.94) (14,0.95)
(16,0.96) (18,0.97) (20,0.98) (22,0.99) (24,1.00)
(26,1.00) (28,1.02) (30,1.03) (32,1.04) (34,1.05)
(36,1.06) (38,1.06) (40,1.07) (42,1.08) (44,1.08)
(46,1.09) (48,1.09) (50,1.10)
};
\addlegendentry{val}

\end{axis}
\end{tikzpicture}
\caption{3DCNN Training and Testing loss after fifty iterations.}
\label{fig:3dcnn-loss}
\end{figure}
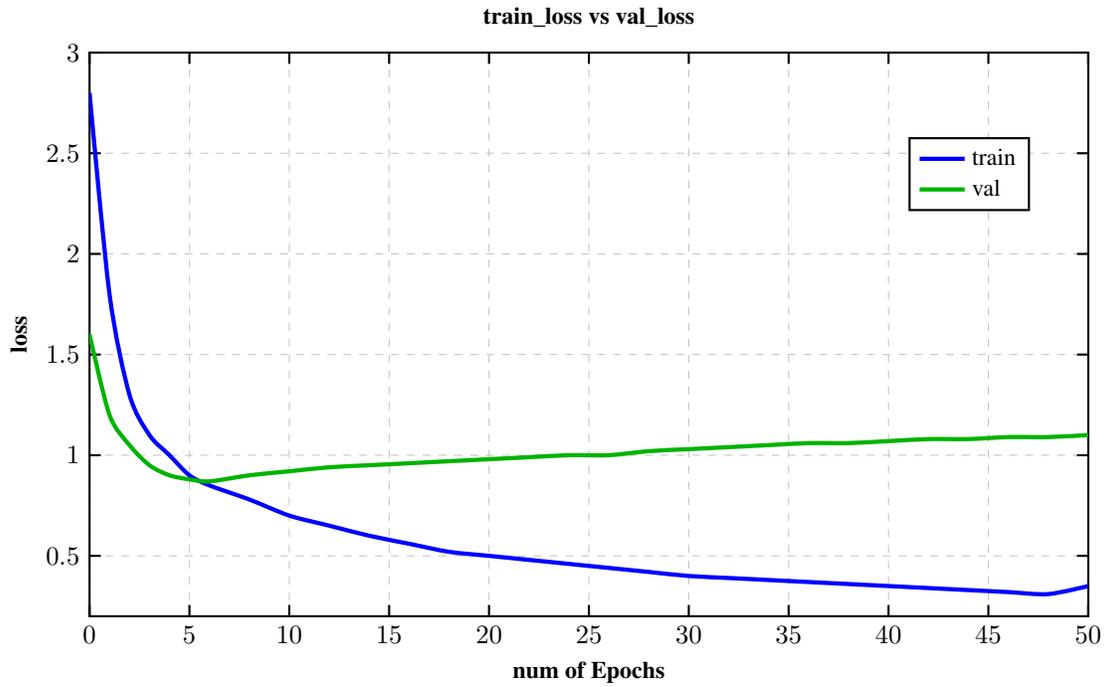

Using 3DCNN, we measured performance for different epoch sizes which helped
to increase the performance of this model. The vowel dataset has
11 classes and 21,783 images data. After training and testing the
vowel dataset using 3DCNN, we got 96.23\% accuracy for training
and 89.90\% accuracy for testing. The consonant dataset has 77,167
images and 39 classes and after training and testing, this model
got a train and test accuracy of 87.23\% and 77.02\%. The numeric
character dataset has 19,748 images data and 10 classes and after
training and testing, we got 98.12\% training accuracy and 94.30\%
testing accuracy. Compound characters i.e. which characters are
the combination of two or more characters have 47,407 images
data and 24 classes and after evaluating the result we got the train
and test accuracy of 92.0\% and 90.0\%. Fig.~\ref{fig:3dcnn-accuracy}
shows the 3DCNN Testing accuracy graph according to dataset type.  
For the whole large dataset, that means a total of 84 classes,
we took 40,000 data for training and testing purposes. After 50
iterations we get the maximum accuracy which is 87.45\% training
accuracy and 76.80\% testing accuracy. Fig.~\ref{fig:3dcnn-train-test}
shows the training and testing accuracy after 50 iterations on the 3DCNN model.
The green line shows the testing accuracy and the blue line shows
training accuracy according to the number of epochs size.  
Fig.~\ref{fig:3dcnn-loss} shows the training and testing loss after 50 iterations,
here the blue line shows the training loss and the green line shows
testing loss according to the number of epochs size.


\paragraph{ResNet50}
\label{sec:res-resnet50}

\begin{table}[htbp]
\centering
\caption{Accuracy of ResNet50 model according to dataset type}
\label{tab:resnet50-accuracy}
\begin{tabular}{|l|c|c|}
\hline
\textbf{Dataset}   & \textbf{Train Accuracy} & \textbf{Test Accuracy} \\
\hline
Vowel              & 96.87\%                 & 93.89\% \\
Consonant          & 96.83\%                 & 93.87\% \\
Numeric            & 97.33\%                 & 96.45\% \\
Compound           & 96.73\%                 & 93.78\% \\
\hline
\end{tabular}
\end{table}

\begin{table}[htbp]
\centering
\caption{Accuracy of ResNet50 model with different epoch sizes}
\label{tab:resnet50-epochs}
\begin{tabular}{|c|c|c|}
\hline
\textbf{Epoch} & \textbf{Train Accuracy} & \textbf{Test Accuracy} \\
\hline
10  & 96.23\% & 92.34\% \\
20  & 96.55\% & 93.57\% \\
30  & 96.88\% & 93.81\% \\
\hline
\end{tabular}
\end{table}


\begin{figure}[htbp]
\centering
\begin{tikzpicture}
\begin{axis}[
    width=0.9\textwidth,
    height=0.55\textwidth,
    xmin=0, xmax=30,
    ymin=0.84, ymax=0.98,
    xlabel={num of Epochs},
    ylabel={accuracy},
    title={train\_acc vs val\_acc},
    grid=both,
    grid style={dashed,gray!40},
    tick style={black, thick},
    label style={font=\bfseries\small},
    title style={font=\bfseries\small},
    legend style={draw=black, fill=white, font=\small, at={(0.85,0.15)}, anchor=south west},
    legend cell align={left},
    thick
]

\addplot+[smooth, ultra thick, color=blue, mark=none]
coordinates {
(0,0.90) (2,0.91) (4,0.92) (6,0.915) (8,0.94) (10,0.93)
(12,0.945) (14,0.955) (16,0.96) (18,0.965) (20,0.97)
(22,0.972) (24,0.975) (26,0.978) (28,0.979) (30,0.98)
};
\addlegendentry{train}

\addplot+[smooth, ultra thick, color=green!70!black, mark=none]
coordinates {
(0,0.86) (2,0.88) (4,0.895) (6,0.90) (8,0.91) (10,0.89)
(12,0.905) (14,0.915) (16,0.92) (18,0.925) (20,0.93)
(22,0.935) (24,0.94) (26,0.945) (28,0.948) (30,0.95)
};
\addlegendentry{val}

\end{axis}
\end{tikzpicture}
\caption{Training and testing accuracy according to epoch sizes.}
\label{fig:resnet-epochs-accuracy}
\end{figure}
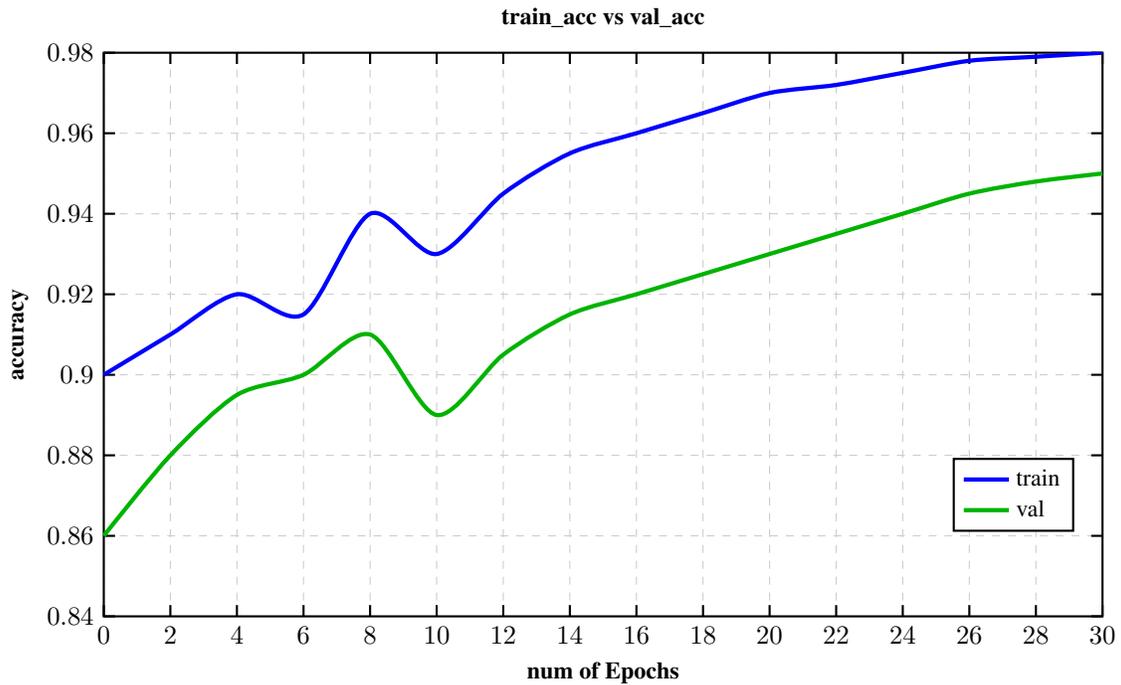


\begin{figure}[htbp]
\centering
\begin{tikzpicture}
\begin{axis}[
    width=0.9\textwidth,
    height=0.55\textwidth,
    xmin=0, xmax=30,
    ymin=0.07, ymax=0.30,
    xlabel={num of Epochs},
    ylabel={loss},
    title={train\_loss vs val\_loss},
    grid=both,
    grid style={dashed,gray!40},
    tick style={black, thick},
    label style={font=\bfseries\small},
    title style={font=\bfseries\small},
    legend style={draw=black, fill=white, font=\small, at={(0.85,0.85)}, anchor=north west},
    legend cell align={left},
    thick
]

\addplot+[smooth, ultra thick, color=blue, mark=none]
coordinates {
(0,0.20) (1,0.185) (2,0.180) (3,0.178) (4,0.176) (5,0.175)
(6,0.172) (7,0.170) (8,0.168) (9,0.165) (10,0.160)
(11,0.170) (12,0.165) (13,0.160) (14,0.158) (15,0.155)
(16,0.150) (17,0.148) (18,0.145) (19,0.140) (20,0.135)
(21,0.130) (22,0.120) (23,0.110) (24,0.100) (25,0.095)
(26,0.090) (27,0.085) (28,0.080) (29,0.078) (30,0.076)
};
\addlegendentry{train}

\addplot+[smooth, ultra thick, color=green!70!black, mark=none]
coordinates {
(0,0.25) (1,0.22) (2,0.21) (3,0.205) (4,0.202) (5,0.198)
(6,0.196) (7,0.197) (8,0.200) (9,0.190) (10,0.195)
(11,0.205) (12,0.192) (13,0.188) (14,0.185) (15,0.180)
(16,0.178) (17,0.170) (18,0.168) (19,0.160) (20,0.155)
(21,0.150) (22,0.145) (23,0.140) (24,0.132) (25,0.125)
(26,0.118) (27,0.110) (28,0.105) (29,0.100) (30,0.098)
};
\addlegendentry{val}

\end{axis}
\end{tikzpicture}
\caption{Training and testing loss according to epoch sizes.}
\label{fig:resnet-epochs-loss}
\end{figure}
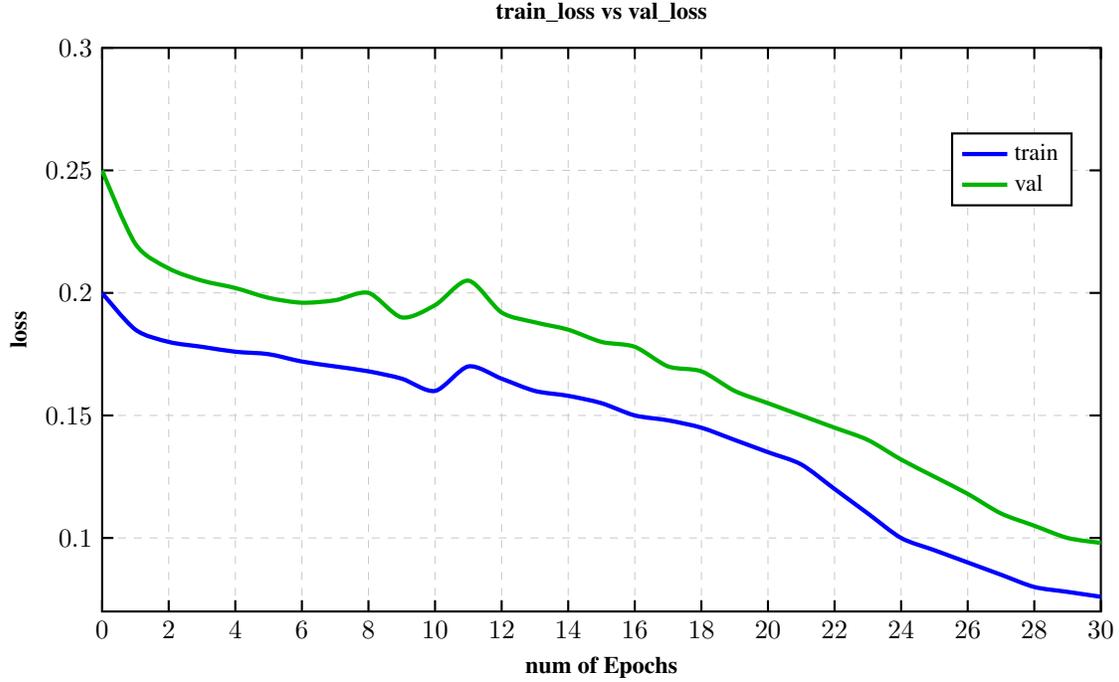

Table~\ref{tab:resnet50-accuracy} shows the training and testing accuracy
for ResNet50. After training and testing the whole dataset using ResNet50 on
different epochs, the train accuracy and test accuracy are presented in
Table~\ref{tab:resnet50-epochs}. The overall train and test accuracies are
96.88\% and 93.81\%, respectively. Fig.~\ref{fig:resnet-epochs-accuracy} shows
the training and testing accuracy for 30 epochs, where the blue line
represents training accuracy and the green line represents testing
accuracy. Fig.~\ref{fig:resnet-epochs-loss} shows the training and testing loss
according to epoch sizes, where the blue line indicates training loss and
the green line indicates testing loss of the model.

\paragraph{MobileNet.}
\label{sec:res-mobilenet}

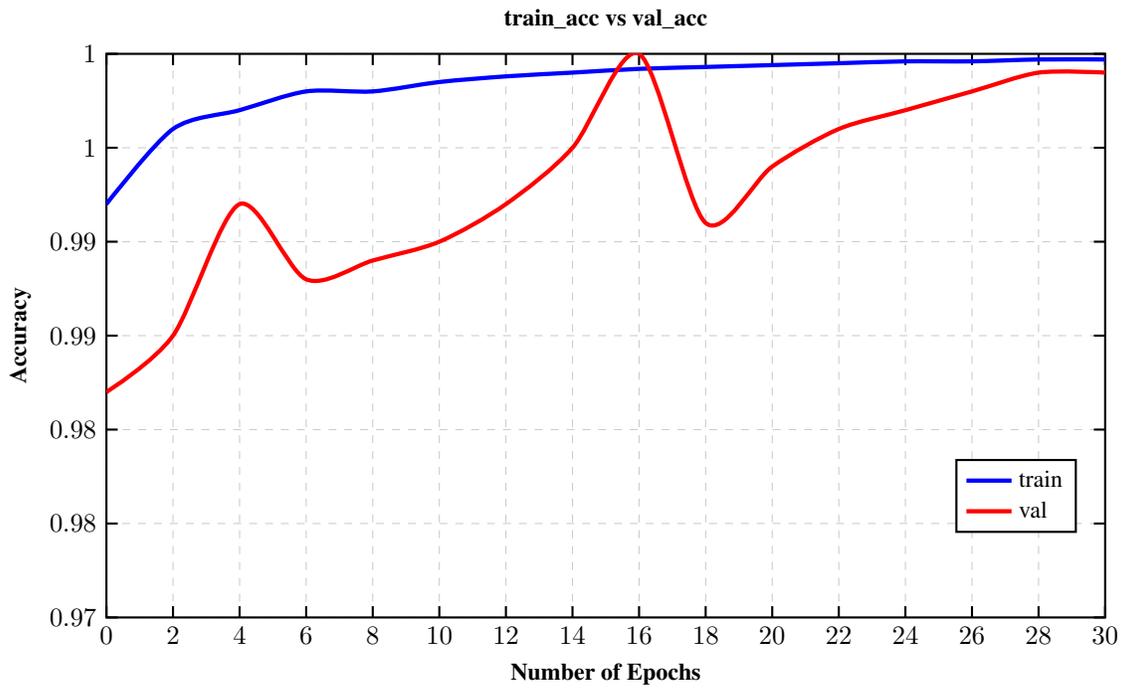
\begin{figure}[htbp]
\centering
\begin{tikzpicture}
\begin{axis}[
    width=0.9\textwidth,
    height=0.55\textwidth,
    xmin=0, xmax=30,
    ymin=0.97, ymax=1.00,
    xlabel={Number of Epochs},
    ylabel={Accuracy},
    title={train\_acc vs val\_acc},
    grid=both,
    grid style={dashed,gray!40},
    tick style={black, thick},
    label style={font=\bfseries\small},
    title style={font=\bfseries\small},
    legend style={draw=black, fill=white, font=\small, at={(0.85,0.15)}, anchor=south west},
    legend cell align={left},
    thick
]

\addplot+[smooth, ultra thick, color=blue, mark=none]
coordinates {
(0,0.992) (2,0.996) (4,0.997) (6,0.998) (8,0.998) 
(10,0.9985) (12,0.9988) (14,0.999) (16,0.9992) 
(18,0.9993) (20,0.9994) (22,0.9995) (24,0.9996) 
(26,0.9996) (28,0.9997) (30,0.9997)
};
\addlegendentry{train}

\addplot+[smooth, ultra thick, color=red, mark=none]
coordinates {
(0,0.982) (2,0.985) (4,0.992) (6,0.988) (8,0.989)
(10,0.990) (12,0.992) (14,0.995) (16,1.000) (18,0.991)
(20,0.994) (22,0.996) (24,0.997) (26,0.998) (28,0.999) (30,0.999)
};
\addlegendentry{val}

\end{axis}
\end{tikzpicture}
\caption{Training and testing accuracy according to epoch sizes.}
\label{fig:mobilenet-epochs-accuracy}
\end{figure}

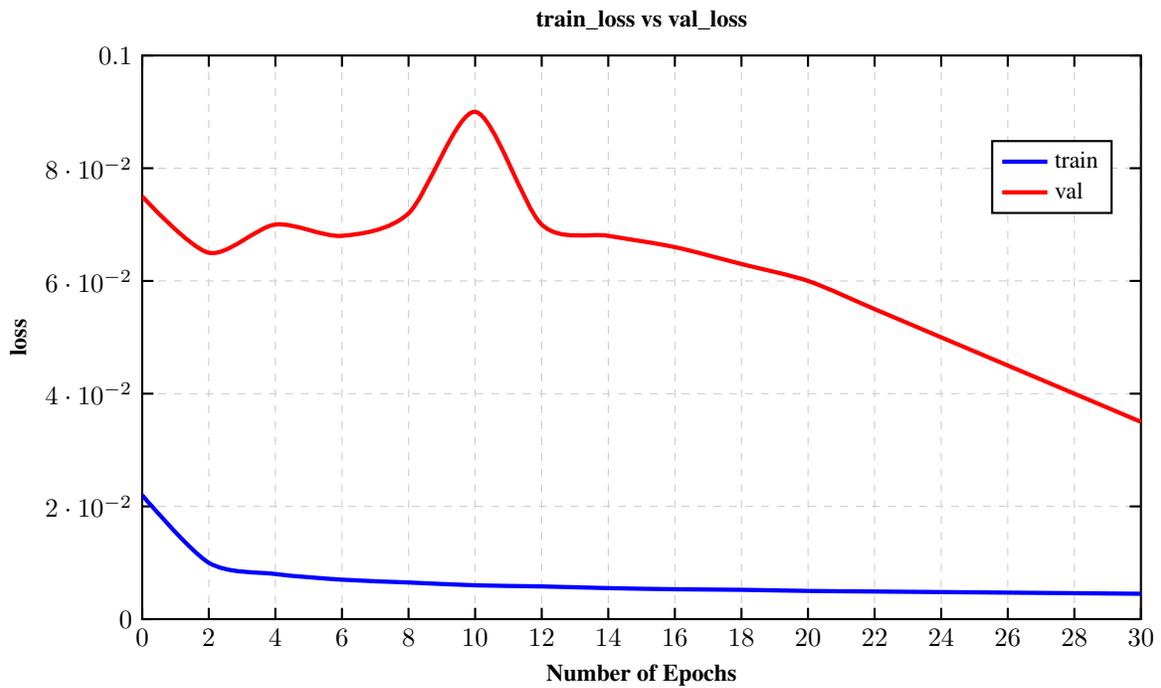
\begin{figure}[htbp]
\centering
\begin{tikzpicture}
\begin{axis}[
    width=0.9\textwidth,
    height=0.55\textwidth,
    xmin=0, xmax=30,
    ymin=0, ymax=0.10,
    xlabel={Number of Epochs},
    ylabel={loss},
    title={train\_loss vs val\_loss},
    grid=both,
    grid style={dashed,gray!40},
    tick style={black, thick},
    label style={font=\bfseries\small},
    title style={font=\bfseries\small},
    legend style={draw=black, fill=white, font=\small, at={(0.85,0.85)}, anchor=north west},
    legend cell align={left},
    thick
]

\addplot+[smooth, ultra thick, color=blue, mark=none]
coordinates {
(0,0.022) (2,0.010) (4,0.008) (6,0.007) (8,0.0065)
(10,0.006) (12,0.0058) (14,0.0055) (16,0.0053) 
(18,0.0052) (20,0.0050) (22,0.0049) (24,0.0048) 
(26,0.0047) (28,0.0046) (30,0.0045)
};
\addlegendentry{train}

\addplot+[smooth, ultra thick, color=red, mark=none]
coordinates {
(0,0.075) (2,0.065) (4,0.070) (6,0.068) (8,0.072)
(10,0.090) (12,0.070) (14,0.068) (16,0.066) (18,0.063)
(20,0.060) (22,0.055) (24,0.050) (26,0.045) (28,0.040)
(30,0.035)
};
\addlegendentry{val}

\end{axis}
\end{tikzpicture}
\caption{Training and testing loss according to epoch sizes.}
\label{fig:mobilenet-epochs-loss}
\end{figure}

Table~\ref{tab:mobilenet-accuracy} shows the MobileNet train and test
accuracy. After training and testing the whole dataset using MobileNet on
different epochs, the training and test accuracies are presented in
Table~\ref{tab:mobilenet-epochs}. The overall train and test accuracies are
99.82\% and 99.46\%, respectively. Fig.~\ref{fig:mobilenet-epochs-accuracy}
shows the training and testing accuracy for 30 epochs, where the blue line
represents training accuracy and the red line represents testing accuracy.
Fig.~\ref{fig:mobilenet-epochs-loss} shows the training and testing loss
according to epoch sizes for the MobileNet model, where the blue line
indicates training loss and the red line indicates testing loss.  

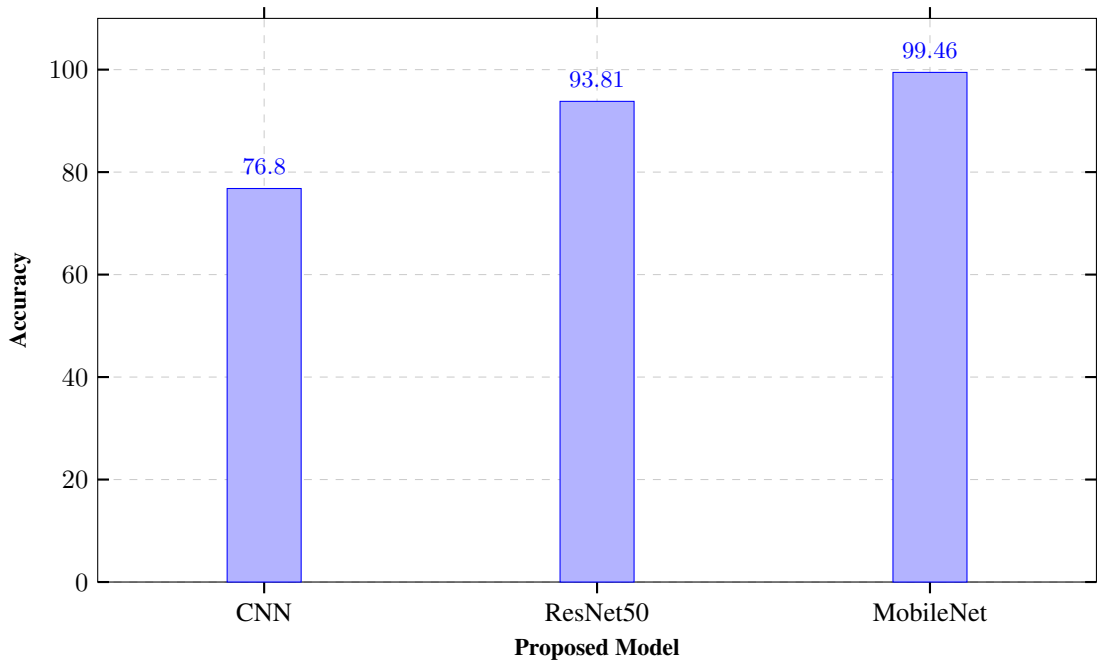
\begin{figure}[htbp]
\centering
\begin{tikzpicture}
\begin{axis}[
    ybar,
    bar width=28pt,
    width=0.9\textwidth,
    height=0.55\textwidth,
    ymin=0, ymax=110,
    ylabel={Accuracy},
    xlabel={Proposed Model},
    symbolic x coords={CNN, ResNet50, MobileNet},
    xtick=data,
    enlarge x limits=0.25,
    grid=major,
    grid style={dashed,gray!40},
    nodes near coords,
    nodes near coords style={font=\bfseries\footnotesize, /pgf/number format/fixed, /pgf/number format/precision=2},
    every node near coord/.append style={yshift=2pt},
    tick style={black, thick},
    label style={font=\bfseries\small},
    title style={font=\bfseries\small},
]
\addplot coordinates {(CNN,76.80) (ResNet50,93.81) (MobileNet,99.46)};
\end{axis}
\end{tikzpicture}
\caption{Our proposed model and accuracy.}
\label{fig:models-comparison}
\end{figure}

Finally, we compare the performance of the three different models.
Using 3DCNN we obtained 76.80\% testing accuracy, with ResNet50 we
achieved 93.81\% testing accuracy, and with MobileNet we reached
99.46\% testing accuracy. Fig.~\ref{fig:models-comparison} shows the accuracy
comparison graph among the three models.

\begin{table}[htbp]
\centering
\caption{Accuracy of MobileNet model according to dataset type}
\label{tab:mobilenet-accuracy}
\begin{tabular}{|l|c|c|}
\hline
\textbf{Dataset}   & \textbf{Train Accuracy} & \textbf{Test Accuracy} \\
\hline
Vowel              & 99.65\% & 99.44\% \\
Consonant          & 99.86\% & 99.55\% \\
Numeric            & 99.87\% & 99.67\% \\
Compound           & 99.70\% & 99.55\% \\
\hline
\end{tabular}
\end{table}

\begin{table}[htbp]
\centering
\caption{Accuracy of MobileNet model with different epoch sizes}
\label{tab:mobilenet-epochs}
\begin{tabular}{|c|c|c|}
\hline
\textbf{Epoch} & \textbf{Train Accuracy} & \textbf{Test Accuracy} \\
\hline
10  & 99.79\% & 99.15\% \\
20  & 99.81\% & 99.29\% \\
30  & 99.82\% & 99.46\% \\
\hline
\end{tabular}
\end{table}


\section{Comparison}
\label{sec:comparison}


\begin{figure}[htbp]
\centering
\begin{tikzpicture}
\begin{axis}[
    ybar,
    bar width=18pt,
    width=0.95\textwidth,
    height=0.55\textwidth,
    ymin=75, ymax=100,
    ylabel={Accuracy},
    xlabel={Model},
    symbolic x coords={
        MLP with Stroke Features,
        MLP and Back Propagation,
        Hidden Markov,
        SVM,
        MLP + SVM,
        ResNet18,
        Our Model
    },
    xtick=data,
    x tick label style={rotate=30, anchor=east, font=\small},
    enlarge x limits=0.1,
    grid=major,
    grid style={dashed,gray!40},
    nodes near coords,
    nodes near coords style={font=\bfseries\footnotesize, /pgf/number format/fixed, /pgf/number format/precision=2},
    every node near coord/.append style={yshift=2pt},
    tick style={black, thick},
    label style={font=\bfseries\small},
    title style={font=\bfseries\small},
]
\addplot coordinates {
    (MLP with Stroke Features,85)
    (MLP and Back Propagation,92)
    (Hidden Markov,88)
    (SVM,89)
    (MLP + SVM,88)
    (ResNet18,95)
    (Our Model,100)
};
\end{axis}
\end{tikzpicture}
\caption{Different models and their accuracy.}
\label{fig:models-accuracy-comparison}
\end{figure}
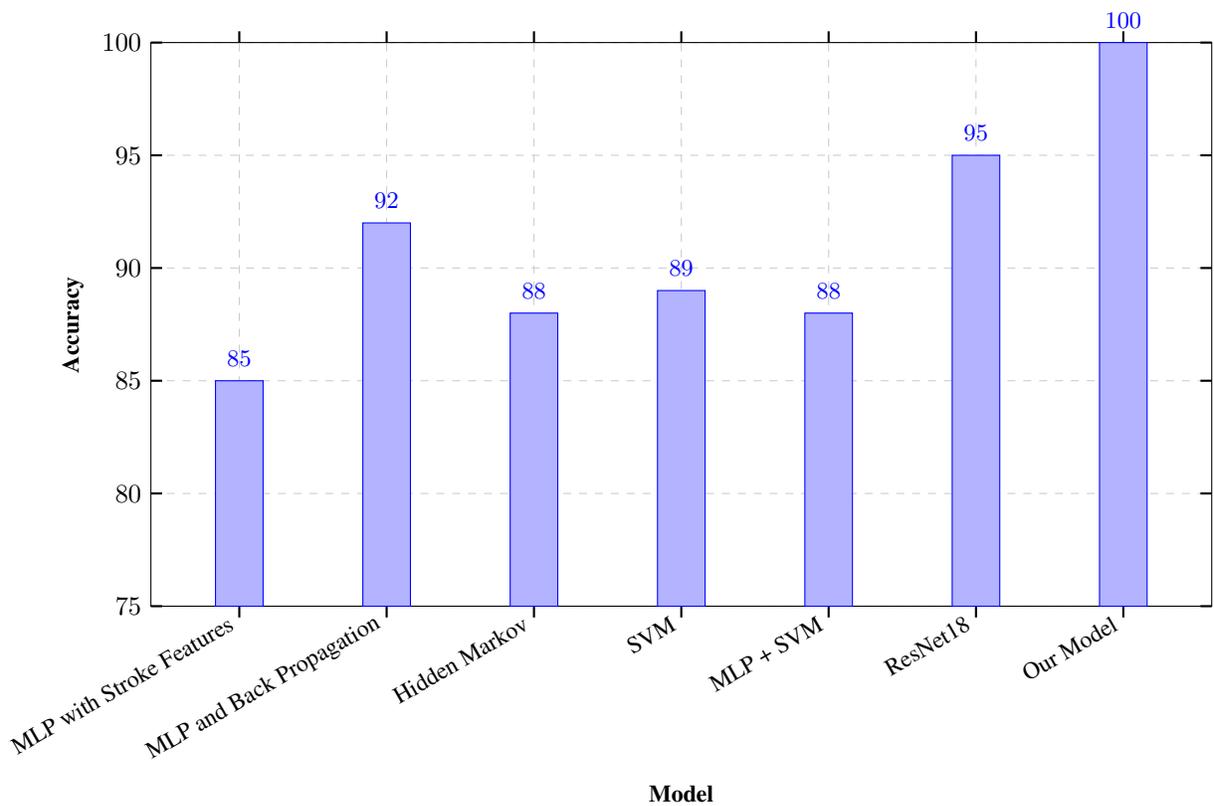

T.~K. Bhowmik and U. Bhattacharya used MLP with Stroke Features, and
achieved 84.33\% accuracy for testing and 89.43\% accuracy for training
\cite{bhowmik2004recognition}. U. Bhattacharya and M. Shridhar used MLP and
Back Propagation, achieving 92.14\% testing accuracy and 94.65\% training
accuracy \cite{bhattacharya2006recognition}. K. Parui and K. Guin applied a
Hidden Markov model and obtained 87.7\% testing accuracy \cite{parui2008online}.
Using SVM-based hierarchical classification, T.~K. Bhowmik and P. Ghanty achieved
89.22\% testing accuracy \cite{bhowmik2009svm}. Sk Mohiuddin and U. Bhattacharya
used MLP and SVM on two different datasets, obtaining 88.79\% and 87.20\% testing
accuracy, respectively \cite{mohiuddin2011unconstrained}. Mujadded Al Rabbani Alif
and Ahmed used a modified ResNet-18 model for the recognition of Bangla isolated
handwriting, reporting 95.10\% testing accuracy \cite{alif2017isolated}.  

Fig.~\ref{fig:models-accuracy-comparison} shows the above-mentioned different
models and the accuracy comparison among them. From this comparison, we can
conclude that our proposed model achieved superior performance compared to the
other models, with a training accuracy of 99.82\% and a testing accuracy of
99.46\%.

\section{Conclusion}
\label{sec:conclusion}

In this paper, we have used transfer learning to classify the basic distinct as well as compound Bangla handwritten characters
while avoiding the vanishing gradient problem. Our proposed method
employs Deep Neural Network techniques such as 3D Convolutional
Neural Network (3DCNN), Residual Neural Network (ResNet), and
MobileNet for the classification of all possible standard formations
of handwritten characters in the Bangla language. Experimental
results demonstrate that our proposed method outperforms other
state-of-the-art algorithms.  

In the future, we aim to upgrade this model so that Bangla handwritten documents can be converted into printed characters with
a high accuracy rate.

\end{document}